\documentclass[conference]{IEEEtran}
\IEEEoverridecommandlockouts

\usepackage{booktabs} 
\usepackage{multirow}

\usepackage[bottom]{footmisc}
\usepackage{blindtext}
\usepackage{graphicx}
\usepackage[font=footnotesize]{caption,subcaption}
\usepackage{wrapfig}

\usepackage{flushend}
\usepackage{amsmath,amsfonts,amsthm,bm}
\usepackage{stfloats}
\usepackage{graphicx}
\usepackage{textcomp}
\usepackage{xcolor}
\usepackage{booktabs} 

\usepackage{amsmath}
\usepackage{rotating}
\usepackage{booktabs}
\usepackage{listings}

\usepackage{listings}
\usepackage{algorithmic}
\usepackage{ragged2e} 
\usepackage{algorithm}
\usepackage[algo2e]{algorithm2e} 

\def\BibTeX{{\rm B\kern-.05em{\sc i\kern-.025em b}\kern-.08em
    T\kern-.1667em\lower.7ex\hbox{E}\kern-.125emX}}
\begin{document}

\title{Saliency Assisted Quantization for Neural Networks \\
}

\author{\IEEEauthorblockN{1\textsuperscript{st} Elmira Mousa Rezabeyk}
\IEEEauthorblockA{\textit{Electrical and Computer Engineering} \\
\textit{Amirkabir University of Technology}\\
Tehran, Iran \\
elmira.rezabeyk@aut.ac.ir}
\and
\IEEEauthorblockN{2\textsuperscript{nd} Salar Beigzad}
\IEEEauthorblockA{\textit{ Software and Engineering} \\
\textit{University of St. Thomas, Minnesota}\\
Minnesota, USA \\
beig2558@stthomad.edu}
\and
\IEEEauthorblockN{3\textsuperscript{rd} Yasin Hamzavi}
\IEEEauthorblockA{\textit{Artificial Intelligence and Robotics} \\
\textit{Isfahan University of Technology}\\
Isfahan, Iran  \\
y.hamzavi@ec.iut.ac.ir}
\and
\IEEEauthorblockN{4\textsuperscript{th} Mohsen Bagheritabar}
\IEEEauthorblockA{\textit{Electrical Engineering} \\
\textit{University of Cincinnati}\\
Ohio, USA \\
baghermn@mail.uc.edu}
\and
\IEEEauthorblockN{5\textsuperscript{th} Seyedeh Sogol Mirikhoozani}
\IEEEauthorblockA{\textit{Electrical and Computer Engineering} \\
\textit{University of Illinois Chicago}\\
Chicago, USA \\
Smirik2@uic.edu}

}
\maketitle

\begin{abstract}
Deep learning methods have established a significant place in image classification. While prior research has focused on enhancing final outcomes, the opaque nature of the decision-making process in these models remains a concern for experts. Additionally, the deployment of these methods can be problematic in resource-limited environments. This paper tackles the inherent black-box nature of these models by providing real-time explanations during the training phase, compelling the model to concentrate on the most distinctive and crucial aspects of the input. Furthermore, we employ established quantization techniques to address resource constraints. To assess the effectiveness of our approach, we explore how quantization influences the interpretability and accuracy of Convolutional Neural Networks through a comparative analysis of saliency maps from standard and quantized models. Quantization is implemented during the training phase using the Parameterized Clipping Activation method, with a focus on the MNIST and FashionMNIST benchmark datasets. 
We evaluated three bit-width configurations (2-bit, 4-bit, and mixed 4/2-bit) to explore the trade-off between efficiency and interpretability, with each configuration designed to highlight varying impacts on saliency map clarity and model accuracy. The results indicate that while quantization is crucial for implementing models on resource-limited devices, it necessitates a trade-off between accuracy and interpretability. Lower bit-widths result in more pronounced reductions in both metrics, highlighting the necessity of meticulous quantization parameter selection in applications where model transparency is paramount. The study underscores the importance of achieving a balance between efficiency and interpretability in the deployment of neural networks.
\end{abstract}


\section{Introduction}
Deep Neural Networks (DNNs) have transformed fields like computer vision, natural language processing, autonomous driving, and healthcare by enabling sophisticated pattern recognition from extensive datasets. These networks have driven notable advancements in accuracy for tasks such as image classification, language translation, and predictive analytics. However, despite their remarkable capabilities, DNNs are frequently criticized for being "black boxes," as their decision-making processes remain largely opaque and difficult to interpret. This opacity presents significant challenges, especially in high-stakes areas like medicine, finance, and autonomous systems, where understanding a model’s decision rationale is essential for building trust, ensuring reliability, and maintaining safety \cite{caruana2015intelligible, li2018tell,goldar2022concept}.
To address these concerns, a significant body of research has focused on developing methods to interpret and explain the decisions made by DNNs. One of the most widely used approaches is the generation of saliency maps, which provide visual representations of the regions in the input data that most strongly influence the model's predictions. Saliency maps are typically produced using gradient-based techniques, which calculate the gradient of the output with respect to the input features to identify areas of importance \cite{selvaraju2017grad, shrikumar2017learning,hassanpour2024overcoming}. These maps are invaluable tools for interpreting model behavior, as they allow users to gain insights into what the model is focused on when making decisions. However, the effectiveness of saliency maps can be compromised by noise and other distracting elements, which may obscure the true regions of interest and lead to misleading interpretations \cite{singh2017hide, kindermans2016investigating}. To mitigate these issues, several advanced techniques have been proposed. For instance, the SmoothGrad method reduces noise by averaging multiple saliency maps generated from slightly perturbed versions of the input \cite{smilkov2017smoothgrad}. Other methods, such as Integrated Gradients \cite{sundararajan2017axiomatic} and Layer-wise Relevance Propagation \cite{bach2015pixel}, modify the backpropagation process to enhance the stability and reliability of the generated saliency maps. Despite these advancements, ensuring that saliency maps are both accurate and robust remains a key challenge, particularly when the model or the input data is subject to slight variations \cite{adebayo2018sanity, ghorbani2019interpretation, bakhshi2024novel}. 
Parallel to the advancement of interpretability methods, there has been growing interest in optimizing neural networks for deployment in resource-limited environments, such as edge devices. A fundamental technique in this optimization is quantization, which reduces the precision of a neural network's parameters and activations, commonly from 32-bit floating-point to lower-precision formats like 8-bit integers. This reduction considerably lowers the memory footprint and computational requirements, making it feasible to deploy complex models on devices with limited resources \cite{Weng2023, karkehabadihlgm, karkehabadi2024ffcl}.

Quantization is essential for the efficient operation of DNNs in settings with constraints on computational power, memory, and energy, such as mobile devices, embedded systems, and other edge computing platforms \cite{Hubara2017, Choi2018}. Various quantization methods have been developed to manage the trade-off between reducing computational complexity and maintaining high accuracy. Quantization-Aware Training (QAT) is particularly effective, as it integrates quantization into the training phase, enabling the network to adjust to lower precision throughout training. This approach generally achieves better accuracy than Post-Training Quantization, which applies quantization to an already trained model without further training \cite{Nagel2019}. Additionally, hardware accelerators like Field-Programmable Gate Arrays (FPGAs) and Application-Specific Integrated Circuits (ASICs) are often used to maximize the performance of quantized models, leveraging their capability to efficiently handle varied numerical representations \cite{Han2016}.
Given the growing importance of both model interpretability and efficient deployment, an emerging question is how these two areas intersect, particularly with regard to saliency. Specifically, as models are increasingly quantized for deployment in resource-limited environments, it becomes imperative to understand how quantization affects the interpretability of these models. \\\\
Does the reduction in precision compromise the quality and reliability of saliency maps? \\
Can quantized models still provide meaningful insights into their decision-making processes, or does quantization obscure these explanations?\\

This paper addresses these critical questions by conducting a comprehensive comparison between "regular model saliency" and "quantized model saliency." We systematically investigate whether the process of quantization affects the generation of saliency maps, and if so, how this impact manifests in different scenarios. Our study applies various saliency methods to both high-precision and quantized versions of neural networks, followed by a detailed analysis of the resulting saliency maps in terms of their clarity, stability, and alignment with the original high-precision models. Through this comparison, we aim to contribute to the broader discourse on balancing model efficiency with interpretability, offering insights that could guide the future development of deployable, yet interpretable, AI systems.

\section{Related Works}

Saliency maps are a widely used tool for interpreting DNNs by highlighting the input features that most influence the model’s predictions. Techniques like Grad-CAM \cite{selvaraju2017grad} and Integrated Gradients \cite{sundararajan2017axiomatic} are popular methods for generating these visual explanations. However, these methods often suffer from noise and instability, leading to challenges in accurately interpreting model behavior \cite{smilkov2017smoothgrad, adebayo2018sanity, pour2024urban}. To mitigate these issues approaches such as SmoothGrad \cite{smilkov2017smoothgrad} and Layer-wise Relevance Propagation (LRP) \cite{bach2015pixel} have been developed, focusing on enhancing the robustness and clarity of saliency maps. In parallel, the rise of neural network quantization has enabled the deployment of DNNs in resource-constrained environments by reducing the precision of model parameters \cite{Weng2023}. Techniques like Quantization-Aware Training \cite{Nagel2019} and mixed-precision quantization \cite{Choi2018} have been crucial in maintaining model accuracy despite reduced computational complexity. However, the impact of quantization on the interpretability of models through saliency maps is not well explored. This paper aims to bridge this gap by comparing "regular model saliency" with "quantized model saliency" to assess how precision reduction affects the generation and reliability of these interpretability tools.

\subsection{Saliency Guided Training}
To address the challenges in explaining the models' behavior, interpretability methods have been developed, with saliency-guided approaches gaining significant attention. Saliency-guided methods focus on identifying and highlighting the most important features in the input data that influence the model's predictions. Gradient-based techniques such as vanilla Gradient \cite{simonyan2013deep}, Grad-CAM \cite{selvaraju2017grad}, and SmoothGrad \cite{smilkov2017smoothgrad} generate saliency maps by computing the gradients of the output with respect to the input, providing a visual representation of the areas most relevant to the model's decisions.

While the post-hoc methods offer valuable insights, they can sometimes produce noisy or unclear saliency maps, highlighting irrelevant features. To overcome these limitations, Saliency Guided Training (SGT) \cite{ismail2021improving} incorporates saliency directly into the training process. This approach teaches the model to focus on important features during training, thereby improving both the accuracy and the interpretability of the resulting saliency maps.

        
        
\begin{algorithm}[H]
\raggedright
 \caption{Saliency Guided Training \cite{ismail2021improving}}
 \label{SGT Alg}
\begin{algorithmic}
 \STATE \textbf{Inputs:} Training samples $X$, number of features to be masked $k$, learning rate $\tau$, hyperparameter $\lambda$
 \STATE Initialize model parameters $f_{\theta}$
 \FOR{$i = 1$ \textbf{to} epochs}
    \STATE \# Obtain sorted index $I$ for the gradient of the output with respect to the input
    \STATE $I = \text{Sort}(\nabla_X f_{\theta_i}(X))$
    
    \STATE \# Mask the bottom $k$ features of the original input
    \STATE $\widetilde{X} = M_k(I, X)$
    
    \STATE \# Compute the loss function
    \STATE $L_i = \mathcal{L}(f_{\theta_i}(X), y) + \lambda \mathcal{D_{KL}}(f_{\theta_i}(X) \| f_{\theta_i}(\widetilde{X}))$
    
    \STATE \# Update network parameters using the gradient
    \STATE $f_{\theta_{i+1}} = f_{\theta_i} - \tau \nabla_{\theta_i} L_i$
 \ENDFOR
\end{algorithmic}
\end{algorithm}


Algorithm \ref{SGT Alg} describes the SGT process, where the model parameters \( \theta \) are iteratively updated based on the importance of input features determined by their gradients. This method encourages the model to focus on the most relevant features during training, improving both interpretability and prediction accuracy. The SGT algorithm employs two key components in the loss function:\\

\begin{itemize}
    \item \textbf{Cross-Entropy Loss}: This widely used loss function quantifies the discrepancy between the model's predicted output \( f_{\theta}(X_i) \) and the actual label \( y_i \). It plays a crucial role in guiding the model to make accurate predictions by adjusting parameters to minimize this difference during training.
    \[
    \text{Cross-Entropy Loss:} \quad \mathcal{L}(f_{\theta}(X_i), y_i)
    \]
    \\
    \item \textbf{Kullback–Leibler (KL) Divergence}: This term quantifies the similarity between the output distributions of the original input \( X_i \) and the masked input \( \widetilde{X}_i \). Minimizing this divergence encourages the model to produce similar outputs even when less important features are masked, ensuring that the model focuses on salient features.
    \[
    D_{KL}(P\|Q) = \sum_{x \in X} P(x) \log\left(\frac{P(x)}{Q(x)}\right)
    \]
\end{itemize}

The overall loss function in SGT combines these components, with a regularization term \( \lambda \) that balances the model's accuracy with its focus on salient features:

\[
\text{Overall Loss:} \quad \sum_{i=1}^{n} \left[\mathcal{L}(f_{\theta}(X_i), y_i) + \lambda \mathcal{D_{KL}}(f_{\theta}(X_i) \| f_{\theta}(\widetilde{X}_i))\right]
\]

During training, the model's parameters are updated to minimize this overall loss, guiding the model to prioritize the most important features and reduce the influence of less relevant ones. By incorporating saliency into the training process, SGT not only enhances model accuracy but also improves the interpretability of the model's predictions, making it particularly valuable in applications where understanding the model's reasoning is as important as the predictions themselves.\\

\subsection{Quantization}

Quantization is a key technique in neural network optimization that reduces the precision of network parameters and activations, typically from high precision (e.g., 32-bit floating point) to lower precision formats (e.g., 8-bit integer). This precision reduction markedly lowers both memory usage and computational demands, making it feasible to deploy sophisticated neural networks on resource-constrained devices, including mobile phones, embedded systems, and other edge devices \cite{Weng2023}.\\\\

\textbf{Quantization-Aware Training (QAT)}

One of the most effective approaches to implementing quantization is QAT. Unlike PTQ, which applies quantization after the model has been fully trained, QAT incorporates quantization directly into the training process. This allows the network to learn and adapt to the constraints imposed by quantization throughout its training, resulting in a model that is more robust and accurate under quantized conditions. During QAT, simulated quantization is applied in both the forward and backward passes. The forward pass uses a simulated quantization function to approximate the behavior of the quantized weights:

\begin{equation}
w_{\text{out}} = (w_{\text{float}}) = \Delta \cdot \text{clamp}\left(0, N_{\text{levels}} - 1, \text{round}\left(\frac{w_{\text{float}}}{\Delta}\right) - z\right)
\end{equation}

Here, \( \Delta \) represents the step size, \( N_{\text{levels}} \) denotes the number of quantization levels, and \( z \) is the zero-point offset used to shift the quantized values.

In the backward pass, the Straight-Through Estimator (STE) is employed to approximate the gradient of the quantization function. This allows the gradients to be propagated through the quantization operation without being blocked:

\begin{equation}
\delta_{\text{out}} = \delta_{\text{in}} \cdot I_{w_{\text{float}} \in (w_{\text{min}}, w_{\text{max}})}
\end{equation}

The weight update during training is then performed using the standard gradient descent approach, but with gradients calculated with respect to the quantized weights:

\begin{equation}
w_{\text{float}} = w_{\text{float}} - \eta \cdot \frac{\partial L}{\partial w_{\text{out}}}
\end{equation}

This methodology significantly narrows the accuracy gap between floating-point and quantized models, particularly at lower bit-widths, as it enables the network to learn to operate within the quantized constraints from the outset of training \cite{Nagel2019b}. Batch Normalization (BN) introduces additional challenges in QAT. During training, BN layers utilize batch statistics (mean and variance), whereas during inference, they rely on running averages. This discrepancy can cause instability in the quantized model. To mitigate this issue, a correction factor is applied during training to better align the batch statistics with the running averages:

\begin{equation}
c = \frac{\sigma_B}{\sigma}
\end{equation}

\begin{equation}
w_{\text{corrected}} = c \cdot \frac{\gamma W}{\sigma_B}
\end{equation}

In these equations, \( \sigma_B \) and \( \sigma \) are the batch and running average standard deviations, respectively, and \( \gamma \) is the scale parameter from batch normalization. Applying this correction ensures that the model remains stable during both the training and inference phases \cite{Ioffe2015}. QAT offers several distinct advantages over PTQ. By integrating quantization into the training process, QAT allows the network to adjust its parameters to better accommodate the quantized representation, often resulting in higher accuracy than PTQ, especially for models that are sensitive to changes in precision \cite{Nagel2019b}.

Furthermore, QAT supports advanced quantization techniques such as per-channel quantization and mixed-precision quantization. Per-channel quantization allows for different channels within a convolutional layer to be quantized independently, which accommodates the varying dynamic ranges across channels. Mixed-precision quantization assigns different bit-widths to different layers based on their sensitivity to quantization, optimizing the trade-off between model size and accuracy \cite{Krishnamoorthi2018}.\\

\textbf{Parameterized Clipping Activation for QAT}\\

To further enhance the performance of quantized neural networks, the PACT method was proposed by \cite{Choi2018}. PACT introduces a learnable clipping parameter during training, addressing the challenge of quantizing activation functions by minimizing quantization error. The clipping function is defined as:

\begin{equation}
a_{\text{clip}} = \text{clip}(a, -\alpha, \alpha)
\end{equation}

In this context, \( \alpha \) is a learnable parameter that determines the clipping threshold. By optimizing this parameter alongside the network weights, PACT dynamically adjusts the clipping threshold based on the training data, leading to significant improvements in the performance of quantized models. In conclusion, quantization, particularly through Quantization-Aware Training, plays a crucial role in enabling the deployment of neural networks in resource-constrained environments. Through techniques like batch normalization correction, per-channel quantization, mixed-precision quantization, and parameterized clipping activation, QAT enhances the accuracy and stability of quantized models, making it an indispensable tool in modern neural network optimization.\\

\subsection{Floating Point Operations (FLOPs) in Quantization}
In the process of quantization, the computational efficiency of a neural network model is significantly impacted by the reduction in bitwidth, as evidenced by the analysis of the model's FLOPs. For the regular model, which operates at a 32-bit precision, the total number of FLOPs is approximately 32 million. However, by quantizing the first and second convolutional layers to lower bitwidths—specifically, 2-bit and 4-bit—the computational load is reduced dramatically. For instance, quantizing both layers to 2 bits reduces the FLOPs to around 3.1 million, representing a tenfold decrease in computational complexity. Similarly, using a 4-bit precision for the first layer and 2-bit for the second results in approximately 3.1 million FLOPs, while a 4-bit precision for both layers yields approximately 5 million FLOPs. This reduction in FLOPs highlights the efficiency gains achieved through quantization, underscoring its potential to optimize model performance for deployment on resource-constrained devices without significantly sacrificing accuracy.

\section{Methodology}

In this study, we compare the saliency maps generated by regular models and quantized models using the PACT method. Our primary focus is on evaluating how quantization affects both the accuracy and the interpretability of CNNs when applied to two benchmark datasets: MNIST and Fashion MNIST.

\subsection{Datasets}

\subsubsection*{MNIST}
The MNIST dataset \cite{lecun2010mnist} comprises 70,000 grayscale images of handwritten digits, each with dimensions of $28 \times 28$ pixels and assigned to one of 10 classes. It is split into a training set of 60,000 images and a test set of 10,000 images.

\subsubsection*{Fashion MNIST}
The Fashion MNIST dataset \cite{xiao2017fashion} includes 70,000 grayscale images of various fashion items, also measuring $28 \times 28$ pixels and divided into 10 classes. This dataset is separated into 60,000 images for training and 10,000 for testing, offering a more challenging benchmark for model evaluation compared to the original MNIST dataset.

\subsection{Model Architectures}

We designed two distinct CNN architectures tailored to the MNIST and Fashion MNIST datasets. Each model includes a straightforward CNN structure with two convolutional layers (kernel size = 3, stride = 1) followed by two fully connected layers. The architectural details of these models are outlined in Table \ref{table:architecture}.

\begin{table}[h!]
\centering
\caption{Model Architectures for MNIST and Fashion MNIST}
\label{table:architecture}
\begin{tabular}{|c|c|}
\hline
\textbf{Layer} & \textbf{MNIST and Fashion MNIST} \\ \hline
\textbf{Conv Layer 1} & 32 filters \\ \hline
\textbf{Conv Layer 2} & 64 filters\\ \hline
\textbf{Fully Connected 1} & 128 neurons \\ \hline
\textbf{Fully Connected 2} & 10 neurons\\ \hline
\end{tabular}
\end{table}

The models were trained using an NVIDIA T4 GPU. For the MNIST dataset, a batch size of 256 and a learning rate of 0.2 were used, while for the FashionMNIST dataset, the training was conducted with a batch size of 128 and a learning rate of 0.1.

\begin{table*}[h!]
\centering
\caption{Accuracy and Flops Results for Regular and Quantized Models}
\label{table:accuracy}
\begin{tabular}{|c|c|c|c|c|}
\hline
\textbf{Dataset} & \textbf{Regular} & \textbf{SGD Quantized (2, 2)} & \textbf{SGD Quantized (4, 2)} & \textbf{SGD Quantized (4, 4)} \\ \hline
\textbf{Accuracy of MNIST}  & 99.46\% & 90.19\% & 98.25\%  & 98.70\%  \\ \hline
\textbf{Accuracy of FashionMNIST} & 93.60\% & 83.86\% & 89.90\% & 90.35\% \\ \hline
\textbf{Number of Flops} & 32,025,344 & 3,108,704 & 3,122,816  & 5,036,480 \\ \hline
\end{tabular}
\end{table*}

\begin{figure*}[h!]
\centering
\includegraphics[width=0.58\textwidth]{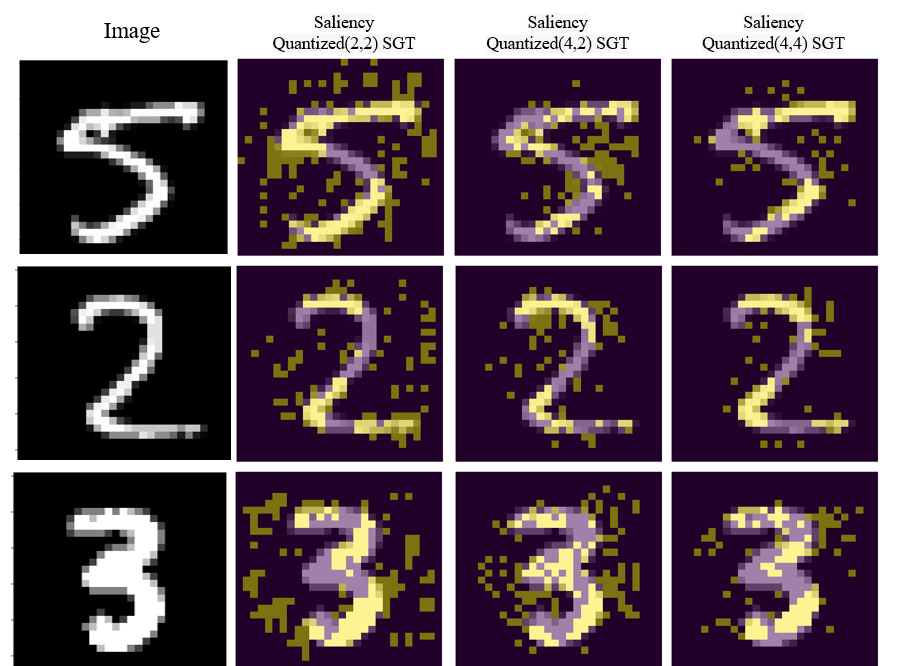}
\caption{Comparison of Saliency Maps for Quantized Models for MNIST dataset}
\label{fig:saliency_maps}
\end{figure*}

\subsection{Quantization and PACT Implementation}
To optimize the CNN models for deployment on resource-constrained devices, we employed quantization using the PACT method. PACT introduces a learnable clipping parameter \(\alpha\) to minimize quantization error in the activation functions, thereby enhancing the performance of quantized neural networks, particularly when lower bit-widths are used \cite{Choi2018}. The clipping function in PACT is defined as:

\[
a_{\text{clip}} = \text{clip}(a, -\alpha, \alpha) = \min\left(\max(a, -\alpha), \alpha\right)
\]

where \(a\) represents the activation values, and \(\alpha\) is a learnable parameter that dynamically adjusts during training to minimize the quantization error. By optimizing \(\alpha\) along with the network weights, PACT effectively reduces the loss in accuracy that typically accompanies quantization. We conducted experiments with different bit-width configurations to evaluate their impact on model accuracy and the quality of saliency maps. The specific bit-width configurations for the layers are summarized in Table \ref{bitwidth}.

\begin{table}[h!]
\centering
\caption{Bit-Width Configurations for Quantized Models}
\label{bitwidth}
\begin{tabular}{|c|c|c|}
\hline
\textbf{Configuration} & \textbf{Layer 1} & \textbf{Layer 2} \\ \hline
\textbf{Configuration 1} & 2 bits & 2 bits  \\ \hline
\textbf{Configuration 2} & 4 bits & 2 bits  \\ \hline
\textbf{Configuration 3} & 4 bits & 4 bits  \\ \hline
\end{tabular}
\end{table}


\begin{figure*}[h!]
\centering
\includegraphics[width=0.6\textwidth]{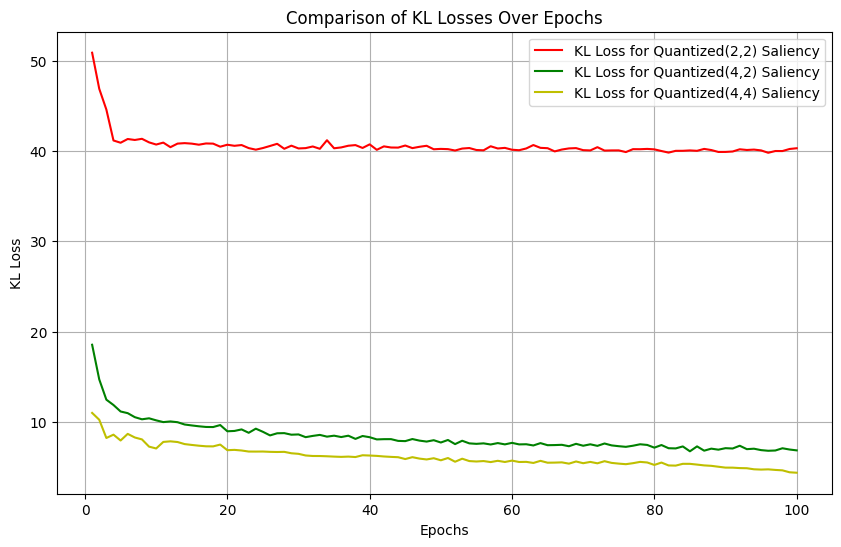}
\caption{Comparison of  KL divergence loss (\(\mathcal{D_{KL}}(f_{\theta_i}(X) \| f_{\theta_i}(\widetilde{X}))\)) for the models on MNIST dataset}
\label{fig:loss}
\end{figure*}
\subsection{Saliency Map Generation}

To generate saliency maps, we utilized the Captum library \cite{kokhlikyan2020captum}, which provides a wide range of tools for model interpretability. We compared the saliency maps produced by the regular models and the quantized models across the different bit-width configurations. The regions of low saliency (low gradient values) were identified and replaced with random values from a uniform distribution for visual comparison.\\

\section{Experiments and Results}

This section presents the results of our experiments, providing a comparative analysis of the quantized models.

\subsection{Accuracy Comparison}

Table \ref{table:accuracy} summarizes the accuracy results for the MNIST and Fashion-MNIST datasets across the different model configurations. As expected, quantization led to a decrease in accuracy, with more significant drops observed in configurations with lower bit-widths. 
Table \ref{table:accuracy} summarizes the accuracy results for the MNIST and Fashion-MNIST datasets across various model configurations. As anticipated, quantization resulted in a decrease in accuracy, with more pronounced drops observed in configurations utilizing lower bit-widths. 
Additionally, Figure \ref{fig:loss} presents the KL divergence loss (\(\mathcal{D_{KL}}(f_{\theta_i}(X) \| f_{\theta_i}(\widetilde{X}))\)) for the models.\\

\begin{figure*}[h!]
    \centering
    \includegraphics[width=0.8\textwidth]{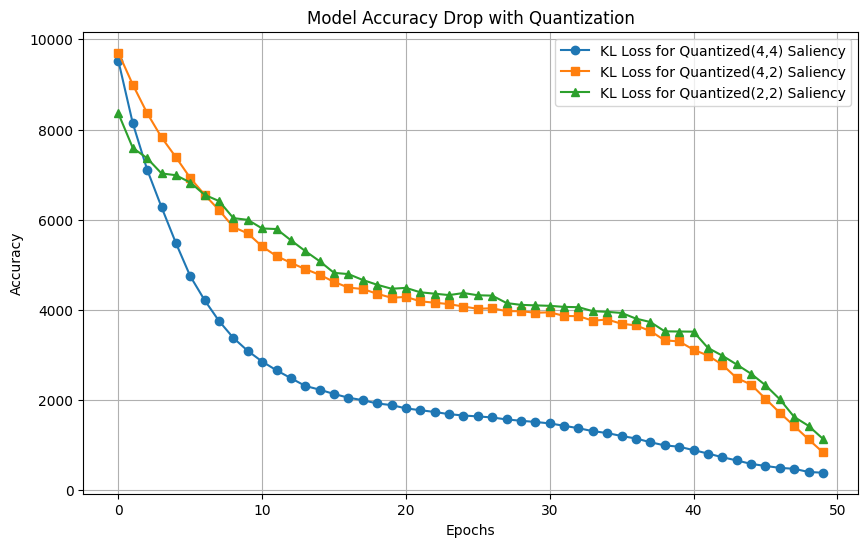}
    \caption{Accuracy drop comparison across different models on MNIST dataset with varying levels of quantization. Although the higher quantization bitwidth (4,4) shows the most salient features as it has a deeper steep. Also, (2, 2) quantization starts with lowest accuracy and drops less which shows lower salient feature learning in the process.  The model with a higher bitwidth(4,4) shows more accuracy drop, indicating better saliency}
    \label{fig:accuracy_drop}
\end{figure*}

\subsection{Saliency Map Comparison}

Figure \ref{fig:saliency_maps} provides a visual comparison of the saliency maps generated by the regular models and the quantized models for both datasets. Saliency maps were generated using gradient-based methods from the Captum library, but it’s important to note that these techniques may be sensitive to noise, which can obscure interpretability, especially in lower bit-width quantizations

As illustrated in Figure \ref{fig:saliency_maps}, the models with higher bits, tend to produce more precise and detailed saliency maps, while those with lower bit-widths, exhibit some degradation in the clarity and sharpness of the saliency maps. This reduction in interpretability aligns with the observed drop in accuracy for these models.

\subsection{Model Accuracy Drop}

In this study, we investigate the effect of quantization on the accuracy of Saliency-Guided Training (SGT) models by leveraging various saliency techniques, including modification-based evaluation \cite{petsiuk2018rise, kindermans2016investigating, karkehabadi2024smoot}. To quantify the impact, we ranked features based on their saliency values and progressively eliminated them, analyzing the resulting changes in model performance. This evaluation was carried out on datasets like MNIST and Fashion-MNIST, which are characterized by well-defined uninformative feature distributions, such as uniform black backgrounds. The results reveal that models subjected to higher levels of quantization exhibit a more pronounced drop in accuracy, irrespective of the saliency technique employed. This finding suggests that lower quantization levels better preserve the model's ability to accurately prioritize and retain essential features, whereas higher quantization levels degrade this capability, leading to sharper decreases in accuracy.

However, it is important to highlight that these conclusions are most applicable to datasets with clearly identifiable uninformative features. The relevance of these findings may diminish when applied to datasets with varying or less predictable background features. In such cases, the inconsistent nature of uninformative features may pose challenges for saliency methods, potentially resulting in less effective feature ranking and elimination, and consequently, different outcomes in model performance. Figure~\ref{fig:accuracy_drop} illustrates the effect of quantization on accuracy drop across different models. As shown, the model with a higher bitwidth (less quantization) experiences a more accuracy drop.




\subsection{Discussion}

The results of our experiments highlight the trade-off between model accuracy and interpretability when applying quantization. While quantization is crucial for deploying models on resource-constrained devices, it inevitably introduces a loss in accuracy and may impact the quality of saliency maps. Our findings underscore the delicate balance between efficiency and interpretability, especially in resource-constrained applications, by demonstrating that even slight modifications to quantization parameters can significantly impact both saliency map clarity and overall model performance.

Future improvements could involve adaptive bit-width configurations that retain interpretability by adjusting parameters for more critical layers or features, thus minimizing accuracy losses, especially when using lower bit-widths. However, as the results suggest, there is still a noticeable degradation in saliency map quality, which is an important consideration for applications where interpretability is critical.

\section{Conclusion}
This study provides an in-depth analysis of the impact of quantization on both the accuracy and interpretability of CNN models. By employing the PACT method and evaluating various bit-width configurations, the research demonstrates that while quantization is essential for optimizing models for resource-constrained environments, it introduces a trade-off between performance and interoperability. Lower bit-widths result in decreased accuracy and less clear saliency maps, particularly in more complex datasets like Fashion MNIST, where precision is vital. The comparison of saliency maps between regular and quantized models reveals that quantization can obscure the model's decision-making process, making the outputs less interpretable. These findings stress the importance of carefully selecting quantization parameters, especially in scenarios where understanding the model's reasoning is as important as its predictions.

In conclusion, while quantization provides significant benefits in reducing computational demands and facilitating the deployment of neural networks on constrained devices, it is crucial to balance these advantages with the potential impact on model interpretability. Future research should aim to develop advanced quantization techniques that preserve interpretability, ensuring that models remain both efficient and transparent in their decision-making processes.

\end{document}